\def\hpec{true}

\ifdefined\plain                % ========== PLAIN

  \ifdefined\review             % ----- draft
    \documentclass[12pt,draft]{article}
    \newcommand{\optsFmt}{format=plain,review}
  \else                         % ----- final
    \documentclass[10pt,final]{article}
    \newcommand{\optsFmt}{format=plain,final}
  \fi

\fi
\ifdefined\hpec                 % ========== HPEC

  \ifdefined\review             % ----- draft
    \documentclass[conference,final]{IEEEtran}
    \newcommand{\optsFmt}{format=hpec,review}
  \else                         % ----- final
    \documentclass[conference,final]{IEEEtran}
    \newcommand{\optsFmt}{format=hpec,final}
  \fi
  
\fi

\usepackage[\optsFmt]{tex/fmtoptions}

\ExplSyntaxOn

\str_case_x:nn {\g_format_str} {
  {plain} {                     % ========== plain
    \usepackage[margin=1in]{geometry}
    \usepackage{authblk}
    \usepackage{setspace}
    \ifreview
      \usepackage{lineno}
      \linenumbers
    \fi }
  {hpec} {                      % ========== HPEC
    \usepackage{balance} }
}

\usepackage{amsmath}
\usepackage{amssymb}
\usepackage{array}
\usepackage{bm}

\usepackage{siunitx}

\usepackage{graphicx}
\usepackage[hyperref,x11names,cmyk,pdftex]{xcolor}

\usepackage[font=footnotesize,hypcap=true]{caption}
\usepackage[font=footnotesize,hypcap=true]{subcaption}

\usepackage{booktabs}
\usepackage{multirow}
\usepackage{dcolumn}
\usepackage{makecell}
\usepackage{adjustbox}

\usepackage{enumerate}
\usepackage[inline,shortlabels]{enumitem}

\usepackage[resetfonts]{cmap}
\usepackage[T1]{fontenc}
\usepackage[utf8]{inputenc}

\usepackage[layout=inline,mode=multiuser,author=]{fixme}

\str_case_x:nn {\g_format_str} {
  {plain} {                     % ========== plain
    \usepackage[pdftex,breaklinks=true,bookmarks=true]{hyperref} }
  {hpec} {                      % ========== HPEC
    \usepackage[pdftex,breaklinks=true,bookmarks=false]{hyperref}
    \usepackage{cite} }
}
\usepackage{cleveref}
\usepackage{textcomp}

\ifdefined\review

  \usepackage[numbers]{natbib}

\fi

\ExplSyntaxOff

\ExplSyntaxOn
\str_case_x:nn {\g_format_str} {
  {hpec} {                      % ========== HPEC
    \IEEEoverridecommandlockouts }
}
\ExplSyntaxOff

\title{
  Spaceland Embedding of Sparse Stochastic Graphs
}

\makeatletter
\let\titletext\@title
\makeatother

\ExplSyntaxOn
\str_case_x:nn {\g_format_str} {
  {plain} {                     % ========== PLAIN
    \ExplSyntaxOff
    \input{tex/authors-plain}
    \ExplSyntaxOn }
  {hpec} {                      % ========== HPEC
    \ExplSyntaxOff
\author{%
  \IEEEauthorblockN{%
    Nikos~Pitsianis\IEEEauthorrefmark{1}\IEEEauthorrefmark{2}   \quad 
    Alexandros-Stavros~Iliopoulos\IEEEauthorrefmark{2}          \quad 
    Dimitris~Floros\IEEEauthorrefmark{1}                        \quad
    Xiaobai~Sun\IEEEauthorrefmark{2}%
  }%
  \\
  \IEEEauthorblockA{%
    \begin{tabular}{c @{\qquad\qquad} c}
      \IEEEauthorrefmark{1}%
      Department~of~Electrical~and~Computer~Engineering
      &
        \IEEEauthorrefmark{2}%
        Department~of~Computer~Science
      \\
      Aristotle~University~of~Thessaloniki
      &
      Duke~University
      \\
      Thessaloniki~54124,~Greece
      &
      Durham,~NC~27708,~USA
    \end{tabular}%
  }%
}

    \ExplSyntaxOn }
}
\ExplSyntaxOff

\newcommand{\datestring}{\today}
\date{\datestring}

\ifreview                       % ----- review mode
  \fxsetup{draft}
\else                           % ----- final mode
  \fxsetup{final}
\fi

\makeatletter
\renewcommand{\@fxlistfixmename}{List of Notes}
\makeatother

\FXRegisterAuthor{ai}{eai}{\color{purple}[AI]}
\FXRegisterAuthor{np}{enp}{\color{green}[NP]}
\FXRegisterAuthor{df}{edf}{\color{blue}[DF]}
\FXRegisterAuthor{xs}{exs}{\color{red}[XS]}

\ifreview                       % ----- review mode
  \newcommand{\pdfauthors}{}
\else                           % ----- final mode
  \newcommand{\pdfauthors}{%
    D. Floros, A.S. Iliopoulos, N. Pitsianis, X. Sun}
\fi

\DeclareSIUnit{\nothing}{\relax}

\newcolumntype{d}[1]{D{.}{.}{#1}}
\newcolumntype{e}[1]{D{e}{\textsc{e}\text{-}}{#1}}
\newcolumntype{m}[1]{D{x}{{\times}}{#1}}
\newcolumntype{x}[1]{D{x}{}{#1}}

\ExplSyntaxOn
\str_case_x:nn {\g_format_str} {
  {plain} {                     % ========== PLAIN
    \crefname{equation}{}{}
  }
  {hpec} {                      % ========== HPEC
    \crefname{equation}{}{}
    \Crefname{equation}{Equation}{Equations}
    \crefname{section}{Section}{Sections}
    \crefname{figure}{Fig.}{Figs.}
  }
}
\ExplSyntaxOff

\hypersetup{%
  pdffitwindow=false,%
  pdfstartview={FitH},%
  colorlinks,%
  pdfauthor={\pdfauthors},%
  pdftitle={\titletext},%
  pagebackref=false,%
  citecolor=PaleGreen4,%
  filecolor=DarkOrchid4,%
  linkcolor=OrangeRed4,%
  urlcolor=black%RoyalBlue4%
}

\setkeys{Gin}{draft=false}

\pdfpageattr{/Group <</S /Transparency /I true /CS /DeviceRGB>>}

\let\leftorig\left
\let\rightorig\right
\renewcommand{\left}{\mathopen{}\mathclose\bgroup\leftorig}
\renewcommand{\right}{\aftergroup\egroup\rightorig}

\makeatletter
\newsavebox{\@tabnotebox}
\providecommand\tmark{} % so having ctable or not is irrelevant
\providecommand\tnote{}
\newenvironment{tabularwithnotes}[3][c]
  {\long\def\@tabnotes{#3}%
   \renewcommand\tmark[1][a]{\makebox[0pt][l]{\textsuperscript{\itshape##1}}}%
   \renewcommand\tnote[2][a]{\textsuperscript{\itshape##1}##2}%\par}
   \begin{lrbox}{\@tabnotebox}
   \begin{tabular}{#2}}
  {\end{tabular}\end{lrbox}%
   \parbox{\wd\@tabnotebox}{
     \usebox{\@tabnotebox}\par
     \smallskip\footnotesize\@tabnotes
   }%
  }
\makeatother

\newcolumntype{d}[1]{D{.}{.}{#1}}

\ExplSyntaxOn
\str_case_x:nn {\g_format_str} {
  {plain} {
    \fxsetup{inline}
    \fxusetheme{color}

  }
}
\ExplSyntaxOff

\hypersetup{final}

\makeatletter
\DeclareRobustCommand\ttfamily{%
  \not@math@alphabet\ttfamily\mathtt
  \fontfamily\ttdefault\selectfont\hyphenchar\font=-1\relax}
\makeatother
\DeclareTextFontCommand{\tturl}{\ttfamily\hyphenchar\font=`/}

\newlist{inparaenum}{enumerate*}{1}
\setlist*[inparaenum,1]{%
  label=(\roman*),
  ref=\roman*,
}
\crefname{inparaenumi}{item}{items}
\Crefname{inparaenumi}{Item}{Items}

\crefname{table}{Table}{Tables}
\Crefname{table}{Table}{Tables}

\crefname{equation}{}{}
\Crefname{equation}{}{}

\graphicspath{%
  {./figs/}%
}

\newcommand{\figwidthBigTwocol}{1.0\linewidth}

\ExplSyntaxOn
\str_case_x:nn {\g_format_str} {

  {plain} {                     % ========== plain

    \newcommand{\figbig}{\figwidthBigOnecol}
  }
  {hpec} {                      % ========== HPEC

    \newcommand{\figbig}{\figwidthBigTwocol}
  }
}
\ExplSyntaxOff

\sfcode`\.=1001
\sfcode`\?=1001
\sfcode`\!=1001
\sfcode`\:=1001
\newcommand{\hmm}[1]{\ifnum\ifhmode\spacefactor\else2000\fi=1001\uppercase{#1}\else#1\fi}

\newcommand{\tsne}{t-SNE}
\newcommand{\tsnebh}{\tsne-BH}
\newcommand{\tsnefi}{FI\tsne}
\newcommand{\tsnepi}{\tsne-$\Pi$}
\newcommand{\sgtsne}{SG-\tsne}
\newcommand{\knn}{$k$NN}

\renewcommand{\vec}[1][x]{\mbs{#1}}
\newcommand{\mat}[1][A]{\mbs{\MakeUppercase{#1}}}
\newcommand{\mbs}[1]{\mathbf{#1}} % math-mode bold symbol

\newcommand{\nNode}{n}
\newcommand{\nEdge}{m}

\newcommand{\bw}[1][i]{\sigma_{#1}}

\newcommand{\Fattr}{\text{attractive term}}
\newcommand{\Frep}{\text{repulsive term}}

\newcommand{\graph}{\mathcal{G}}
\newcommand{\vertices}{V}
\newcommand{\edges}{E}

\newcommand{\weightMat}{\mathbf{P}}
\newcommand{\weightMatCond}{\weightMat_\text{c}}
\newcommand{\weightMatEmb}{\mathbf{Q}}

\newcommand{\setX}{\mathcal{X}}
\newcommand{\setY}{\mathcal{Y}}

\DeclareMathOperator{\dist}{d}
\DeclareMathOperator{\opKL}{KL}
\newcommand{\kl}[2]{\opKL\left( #1 \| #2 \right)}

\begin{document}

%%% =============
%%% FRONT MATTER

\ExplSyntaxOn
\str_case_x:nn {\g_format_str} {
  {plain} {                     % ========== PLAIN
    \ExplSyntaxOff
    \input{tex/frontmatter-plain}
    \ExplSyntaxOn }
  {hpec} {                      % ========== HPEC
    \ExplSyntaxOff
    % frontmatter-hpec.tex
%
% Drafted by Dimitris F. on April 24, 2019

% title
\maketitle

% list of notes (DRAFT only)
\ifreview
\phantomsection
  \pdfbookmark[1]{TODO list}{sec:todo}
  \listoffixmes
\fi

% abstract
\begin{abstract}
  % abstract.tex
%
% Drafted by Dimitris F. on April 24, 2019
%
% Revised by xiaobai on May 24, 2019
% Editing by Alexandros (May 24, 2019)
%
% Revised by Dimitris F. on May 25, 2019
%
%
We introduce a nonlinear method for directly embedding large, sparse,
stochastic graphs into low-dimensional spaces, without requiring
vertex features to reside in, or be transformed into, a metric space.
Graph data and models are prevalent in real-world applications.
Direct graph embedding is fundamental to many graph analysis tasks, in
addition to graph visualization.
We name the novel approach \sgtsne, as it is inspired by and builds upon
the core principle of \tsne, a widely used method for nonlinear
dimensionality reduction and data visualization.
We also introduce t-SNE-$\bm{\Pi}$, a high-performance software for
2D, 3D embedding of large sparse graphs on personal computers with
superior efficiency. It empowers \sgtsne{} with modern computing
techniques for exploiting in tandem both matrix structures and memory
architectures.
We present elucidating embedding results on one synthetic graph and
four real-world networks. More experimental results and comparisons
are in Supplementary Material.\footnote{Supplementary Material is at
  \url{http://t-sne-pi.cs.duke.edu}.}

%
%
%
%%% Local Variables:
%%% mode: latex
%%% TeX-master: "../HPEC2019-TSNEPI-main"
%%% End:

\end{abstract}

%%% Local Variables:
%%% mode: latex
%%% TeX-master: "../HPEC2019-TSNEPI-main"
%%% End:

    \ExplSyntaxOn }
}
\ExplSyntaxOff

%%% ==================
%%% DOCUMENT BODY

% BODY-main.tex
%
% Drafted by Dimitris F. on April 24, 2019
%
% Revised by Xiaobai on May 19-24, 2019
%

% ==================== INTRODUCTION
%

\section{Introduction}
\label{sec:introduction}

% ---------- Introduction (graph embedding}

%  SG-introduction.tex
%
%  by Xiaobai Sun, May 24, 2019
%
% Revised by Dimitris F. on May 25, 2019

Big and sparse graph or network data exist and emerge in various research
fields and real-world applications. Relational data of this type are in
great diversity, such as biological networks, social, friend or co-author
networks, commercial product networks, food webs, ecological networks,
telecommunication networks, information networks, transportation networks,
word co-occurrence networks, image-net,
word-net~\cite{albert1999internet,girvan2002community,%
  kunegis2013konect,Kwak10www,%
  yang2015defining,icwsm10cha,newman2006finding,%
  auer2007dbpedia,yu2008high,redner2004citation,%
  zachary1977information,watts1998collective,%
  Russakovsky2015,Miller1995,Fellbaum1998}.  Graph/network analysis plays
an important role in data analysis in general.

A graph ${\cal G}(V,E)$ has a set $V$ of vertices and a set $E$ of
edges. The vertices (nodes) are an abstraction of entities or objects that
have concrete forms or possess particular attributes in a real-world
context. The vertices may stand for molecules, proteins, neuron cells,
species, products, customers, words, documents, signals or time series,
images or pixels. An edge (link) connecting two vertices represents a certain
relationship, an interaction, or proximity between them.  Additional graph
information may include vertex attributes and edge weights.
% 
% Not to remove; for later use by xiaobai. 
% Graph analysis makes inquiries about and inference
% of global, regional or functional information via propagating and
% aggregating local information at vertices and edges, and reversely,
% makes estimation or prediction on latent vertex properties, or on
% missing/erroneous/unseen edges, from regional, global or functional
% structures and statistics.

Fundamental to many graph analysis tasks, graph embedding renders a
mapping from vertices to their coded vectors (abstract features) in a
code space (the embedding space), with code length $L$. The mapping is
subject to one or more conditions to preserve (or reconstruct) certain
graph properties in the embedding space. If the edge weights indicate
pairwise (geometric or geodesic) distances between nodes, one wishes
to preserve the pairwise distances as much as possible.  This type of
graph embedding is vertex embedding. Often, the code space for large
graphs resides in $\mathbb{R}^{L}$, such as with {\tt word2vec} and
its variants~\cite{Mikolov2013,Pennington2014}.
Various graph analysis tasks are subsequently carried out in an
embedding space. They include (semi-, un-) supervised classification
or stratification, abnormality detection, noise reduction, propagation
patterns, content recommendation and other types of prediction. The
feature vector length $L$ is chosen sufficiently large for these
multiple analytical tasks.
In recent years, numerous types of high-dimensional graph embedding
techniques are tried out with neural
networks~\cite{kramer1991nonlinear,fritzke1995growing,%
  hinton2006reducing,wang2014generalized,
  Tang2015,Levine2015,%
  Klein2015,Tirosh2016,Becher2014,Bronstein2017,Cho2014}.
%

% ============================== FIGURE: MOBIUS
% 
% Mobius-lattice-embedding.tex
% 
% figure prepared by Dimitris F. between May 07-20, 2019
% 
% text and caption revised by Xiaobai Sun on May 19, 2019
% clutters removed by Xiaobai Sun
%
% post-submission revision by xiaobai on May 27, 2019
% 

% ==================== FIGURE: MOBIUS STRIP

\begin{figure}
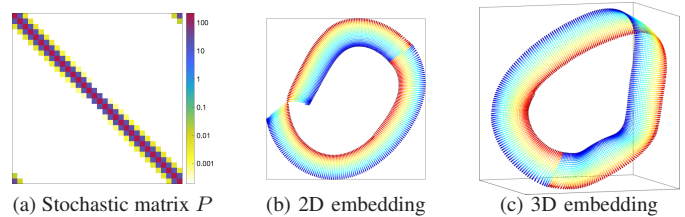

  \centering
  \newcommand{\vskipRow}{0.4em}
  \newcommand{\vskipProj}{0.2em}
  \newcommand{\figWidth}{0.30\linewidth}
  \newcommand{\basepath}{mobius}
  \newcommand{\mobiusGraph}{graph-mobius_szBlk-256}
  \newcommand{\embed}[1]{mobius_lambda-100.0_nn-150_d-#1_uniform-1_maxIter-1000_earlyExag-250_seed-0}
  \begin{subfigure}{\figWidth}
    \centering
    \includegraphics[width=\linewidth]{%
      {{\basepath/\mobiusGraph}}}
  \end{subfigure}
  \hspace*{\fill}
  \begin{subfigure}{\figWidth}
    \centering
    \includegraphics[width=.8\linewidth]{%
      {{\basepath/\embed{2}}}}
  \end{subfigure}
  \hspace*{\fill}
  \begin{subfigure}{\figWidth}
    \centering
    \includegraphics[width=\linewidth]{%
      {{\basepath/\embed{3}}}}
  \end{subfigure}
  % 
  % 
  % --- CAPTIONS
  % 
  \begin{subfigure}{\figWidth}
    \centering
    \caption{Stochastic matrix $P$}
  \end{subfigure}
  \hspace*{\fill}
  \begin{subfigure}{\figWidth}
    \centering
    \caption{2D embedding}
  \end{subfigure}
  \hspace*{\fill}
  \begin{subfigure}{\figWidth}
    \centering
    \caption{3D embedding}
  \end{subfigure}
  \caption[Synthetic data]{\footnotesize %
    Essential neighbor and structural connections in a $k$NN graph for
    a M\"obius strip lattice,  with $n = 8{,}192\ ( =256\!\times\! 32)$
    lattice nodes, are obscured or crumbled by a 2D embedding yet
    principally captured by a 3D  embedding. 
    {\bf (a)}
    The stochastic matrix $P$ of the $k$NN graph, $k=150$,
    by Euclidean distance, is displayed in a  $32\!\times\! 32$ pixel
    array. Each pixel represents a  $256\!\times\! 256$ 
    matrix block, its value (color coded) is the total sum of the block
    elements.
    {\bf (b)\&(c)}
    The embeddings are by \sgtsne\ with 
    $ \lambda = 100 $, taking $1{,}000 $ iterations,
    including $250$ early exaggeration ones with $\alpha = 12$.
    The initial coordinates are drawn from a random uniform distribution,
    The embedding of the same graph by the conventional \tsne\ within
    the feasible perplexity range suffers more distortions, see
    Supplementary Material.
  } % end of caption 
  \label{fig:embedding-with-synthetic-data}
\end{figure}

%%% Local Variables:
%%% mode: latex
%%% TeX-master: "../HPEC2019-TSNEPI-main"
%%% End:

%
% ===============================================

High-dimensional graph embedding is often followed and accompanied by
dimension reduction and low-dimensional embedding for multiple purposes:
\begin{inparaenum}[(i)]
\item visual assessment, inspection and summary of the analysis
  results in large quantities;
\item facilitating interactive exploration; and 
\item making connections to the original, interpretable attributes.
\end{inparaenum}
There is a plethora of low-dimensional graph embedding
techniques~\cite{Mead1992,fua1999hierarchical,%
  huang2005multidimensional,mcinnes2018umap}.
Most are distance-based, preserving geometric neighborhood, assuming
vertex features in a metric space.

We make two major contributions in the present work. First, we
introduce a novel nonlinear approach for directly embedding large,
sparse, stochastic graphs into low-dimensional spaces, without
requiring vertex features to reside in, or be transformed into, a
metric space.
There are a lot of stochastic graphs, in real-world applications,
without feature vectors readily in a metric space or available for
public use.
Our approach, named \sgtsne, is inspired by and builds upon the core
principle of \tsne, which is widely used for nonlinear dimensionality
reduction and data visualization. There are many variants of
t-SNE~\cite{Maaten2008,Amir2013,VanDerMaaten2014,Abdelmoula2016,%
  Pezzotti2016,Pezzotti2017,Unen2017,Linderman2019}.
In \sgtsne, by a radical removal of the existing restrictions, we extend
the use of t-SNE to the entire realm of stochastic graphs, no longer
limited to distance-based $k$-nearest neighbor ($k$NN) graphs.  \sgtsne\ is
also equipped with a parameterized non-linear rescaling mechanism.  It
explores the sparsity to a much greater scope. In the case of a $k$NN
graph, \sgtsne\ is capable of rendering much better embedding results than
conventional t-SNE. See experimental comparisons in Supplementary Material.
% at \url{http://t-sne-pi.cs.duke.edu}.

Secondly, we introduce a software \tsnepi{}.  It accelerates 2D graph
embedding up to $5\times$ in execution time in comparison to the best
among the existing t-SNE variants. More importantly, it enables 3D
embedding of large sparse graphs on personal computers, with superior
efficiency.
We demonstrate in \cref{fig:embedding-with-synthetic-data} that 3D
embedding has greater capacity of preserving local and structural
connectivity.
Steerable visualization of all 3D embedding results is available at the
online Supplementary Material.
% 
% We make use of advanced computing knowledge and techniques to mitigate
% typical problems in data access latency when sparse data are large and
% irregularly scattered.
%
We present also experimental results on two biological networks, a
social network created by Google, and an Amazon product network. The
embedding results are characteristic and elucidating.

% They manifest multiple advantages of the novel method \sgtsne{} with
% high-performance support by \tsnepi{}.

%%% End:

%%% Local Variables:
%%% mode: latex
%%% TeX-master: "../HPEC2019-TSNEPI-main"
%%% End:

\section{Embedding sparse stochastic graphs}
\label{sec:SG-abstraction}

% review-of-tsne.tex
%
% drafted by Xiaobai Sun on May 19, 2019
% edited by Alexandros (May 24, 2019)
%
% Revised by Dimitris F. on May 25, 2019

We introduce \sgtsne{}, with which we apply and extend the essential
principle of \tsne\ to the entire realm of sparse stochastic graphs.
A stochastic graph $\graph = (\vertices, \edges, \weightMatCond)$ is
associated with a stochastic weight matrix
$\weightMatCond = [p_{j|i}]$.  At each vertex $i$,
$ \sum_{j} a_{ij} \, p_{j|i} =1$, where $\mathbf{A} = [a_{ij}] $ is
the binary-valued adjacency matrix (the sparsity mask) of the graph.
% the conditional neighbor probability distribution
We assume without loss of generality that $\graph$ has no isolated
($0$-degree) vertices.  We are especially concerned with real-world
graphs that have a large number of vertices and sparse connections
($ |E| = O( |V| $). Real-world graphs do not necessarily have constant
degrees.

% ==================== FIGURE: PBMC-8K EMBEDDING
% 
% fig-amazon.tex
%
% Drafted by Dimitris F. on May 23, 2019

% ============================== FIGURE: AMAZON
% 
\begin{figure}
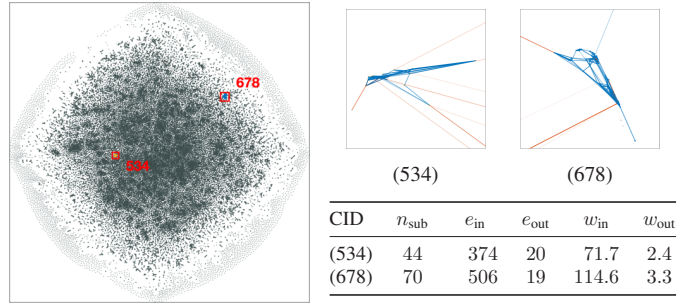

  \centering
  % ~~~~~ AMAZON
  % 
  % --- renew figure placement options
  % 
  % space between sub-communities on the same line
  \newcommand{\sepHrzFig}{\quad}
  % 
  % width of sub-community figures
  \newcommand{\subFigWdt}{.25\linewidth}
  % 
  % --- commands to load data
  \newcommand{\basepath}{graph-embedding/amazon}
  \newcommand{\amazonTot}
  {Amazon_lambda-10.0_d-2_gaussian-1.000000e-04_maxIter-8000_earlyExag-2000_seed-0}
  \newcommand{\amazonSub}[1]{\amazonTot_with-edges_comm-#1}
  % 
  % --- 
  % 
  % -------------------- EMBEDDING
  % 
  \begin{subfigure}{\linewidth}
    \centering
    \begin{subfigure}{.45\linewidth}
      \includegraphics[width=\linewidth]{%
        {{\basepath/\amazonTot}}}
    \end{subfigure}
    \hspace*{\fill}
    % 
    % -------------------- COMMUNITY DETAILS
    % 
    \begin{subfigure}{.52\linewidth}
      \centering
      %
      % ---------- orange sub-graph
      %
      \begin{subfigure}{.4\linewidth}
        \includegraphics[width=\linewidth]{%
          {{\basepath/\amazonSub{534}}}}
        \caption*{(534)}
      \end{subfigure}
      \quad
      % 
      % ---------- blue sub-graph
      % 
      \begin{subfigure}{.4\linewidth}
        \includegraphics[width=\linewidth]{%
          {{\basepath/\amazonSub{678}}}}
        \caption*{(678)}
      \end{subfigure}
      \\[0.5em]
      % 
      % ---------- edge statistics
      % 
      \begin{subfigure}{\linewidth}
        \centering
        \begin{adjustbox}{max width=\linewidth}
          \begin{tabular}{ @{}
            l
            S[table-format=2]
            S[table-format=4]
            S[table-format=2]
            S[table-format=3.1]
            S[table-format=1.1]
            @{} }
            \toprule
            \multicolumn{1}{@{}l}{CID}
            & \multicolumn{1}{c}{$n_{\text{sub}}$}
            & \multicolumn{1}{c}{$e_{\text{in}}$}
            & \multicolumn{1}{c}{$e_{\text{out}}$}
            & \multicolumn{1}{c}{$w_{\text{in}}$}
            & \multicolumn{1}{c@{}}{$w_{\text{out}}$}
            \\
            \midrule
            (534)  & 44 & 374  & 20   & 71.7 & 2.4    \\ 
            (678)  & 70 & 506  & 19   & 114.6& 3.3    \\   
            \bottomrule
          \end{tabular}
        \end{adjustbox}
      \end{subfigure}
    \end{subfigure}
    
  \end{subfigure}
  % 
  % -------------------- CAPTION -------------------------------- 
  % 
  \caption[Amazon embedding]{%
    Two-dimensional embedding of an \texttt{Amazon} product network,
    with $\nNode = \num{334 863}$ products and $\nEdge = \num{1 851 744}$
    edges, each edge represents connects two products that are
    frequently purchased together~\cite{Yang2015}.
    \textbf{Left:}~Vertex embedding. The highlighted vertices
    are two disjoint product clusters with labels
    $534$ and $678$.  The embedding is enabled by \sgtsne{}, 
    with $\lambda = 10$, taking $8{,}000$ iterations, including
    $2{,}000$ early exaggeration iterations with
    $\alpha = 12$. The initial coordinates are
    drawn from a random Gaussian distribution.
    The graph is inadmissible to the conventional \tsne{}.
    \textbf{Right:}~The subgraphs of the two clusters with
    intra-edges in blue (close together) and external edges in red.
    The table details each of the clusters in numbers of nodes,
    internal edges and external edges, and the sum of weighs
    over the internal and external edges.
    More scatterplots of the network are available in Supplementary
    Material.  }
  \label{fig:amazon}
\end{figure}

%%% Local Variables:
%%% mode: latex
%%% TeX-master: "../HPEC2019-TSNEPI-main"
%%% End:

%
% ==============================================

\subsection{An investigative review of \tsne}

% ======== the essence

We describe first the essence of \tsne. Provided with a graph
${\cal G}$ with stochastic weights at each vertex, described by a
stochastic matrix $\weightMatCond = (P_{j|i})$, and given a prescribed
modest dimensionality $d$, \tsne\ generates vertex embedding
coordinates $\setY_{d}^{\ast} = \{ \vec[y] \}_{i=1}^n$ in
$\mathbb{R}^d$.  With its configuration specified by the Student
t-distribution $\weightMatEmb(\setY) = [q_{ij}]$,
\begin{equation}
\label{eq:Q-elements} 
  q_{ij} = \frac{ (1 + \| \mathbf{y}_i - \mathbf{y}_j \|^2)^{-1} } 
         {\sum_{k \neq l}(1 + \| \mathbf{y}_k - \mathbf{y}_l \|^2)^{-1}} ,
         \quad 
         % \forall (i,j),
         \ i\neq j ,
\end{equation}
the embedding arrives at the optimal matching to the joint
distribution
$ \weightMat = ( \weightMatCond + \weightMatCond^{\mathsf{T}} ) / 2n $
in the sense that
\begin{equation}
  \label{eq:distribution-matching-kl}
  \setY_{d}^{\ast}  = \arg \min_{\setY}
  \kl{ \weightMat }{ \weightMatEmb(\setY) } ,
\end{equation}
where KL denotes the Kullback-Leibler divergence.
%
% In machine learning terminology, $\setY$ is the set of parameters to be
% determined.

% === standard version 

The standard version of \tsne{}, however, is developed, used, viewed
and reviewed as a nonlinear dimension reduction algorithm, due
primarily to the following interface to the distribution matching
objective \cref{eq:distribution-matching-kl}.
The input is a set of data points in terms of feature vectors
$\setX = \{ \vec[x]_{i} \}_{i=1}^n$ in an $L$-dimensional space
$\mathbb{R}^{L}$ endowed with a distance function
$\dist(\cdot, \cdot)$.  The feature vector length $L$ is (much) larger
than the embedding dimension $d$.
The algorithm locates the $k$ nearest neighbors ($k$NN) of each vertex, and
integrates them into the $k$NN graph, which is regular (with constant degree
$k$).  Let $\mathbf{A}$ be the binary-valued adjacency matrix.
Denote by $\mathbf{D}$ the matrix of squared pair-wise distances,
$\mathbf{D} = [ d_{ij}^2 ] = [\dist^2( \mathbf{x}_i,\mathbf{x}_j)] $. Then,
the distance-weighted $k$NN matrix $\mathbf{D}\odot \mathbf{A}$ (Hadamard
product) is converted to a stochastic matrix $\weightMatCond$ by
parameterized exponentiation and normalization,
\begin{equation}
\label{eq:d-to-p}
  p_{j|i}(\sigma_i)  = 
    \frac{ a_{ij} \exp\left(-d_{ij}^2 / 2 \bw^2 \right) }
    { \sum_{l}
      a_{il} \exp\left(-d_{il}^2 / 2 \bw[i]^2 \right) }.
\end{equation}
For every vertex $i$, the Gaussian parameter $\sigma_{i}$ is
determined by the equation
\begin{equation}
\label{eq:sigma-equalization}
  - \sum_{j} a_{ij}  p_{j|i}(\bw)  \log(   p_{j|i}(\bw) ) = \log(u), 
  % \quad \forall i \in \{1, \ldots, n\} ,
\end{equation}
where the prescribed perplexity $u$ is a hyper-parameter. In fact,
equation~(\ref{eq:sigma-equalization}) equalizes all vertices by the
conditional entropies.

% ==================================== figure ORKUT 
% 
% figure-orku-2D.tex
%
% created by Dimitris F.
%
% caption revised by xioabai May 24-25, 2019
% 

% ==================== FIGURE: 2-D EMBEDDING OF ORKUT

\begin{figure*}
  \centering
  % --- options for figure placement
  %
  % space between sub-communities on the same line
  \newcommand{\sepHrzFig}{\hspace*{0.3em}}
  % 
  % space between sub-communities lines
  \newcommand{\sepVrtFig}{0.3em}
  %
  % width of sub-community figures
  \newcommand{\subFigWdt}{.30\linewidth}
  % \newcommand{\subFigWdt}{.21\linewidth}
  %
  % width of histogram
  \newcommand{\histWdt}{.95\linewidth}
  % \newcommand{\histWdt}{.65\linewidth}
  %
  % separation of thumbnails and histogram
  \newcommand{\sepVrtHist}{0.3em}
  % 
  % --- commands to load data
  \newcommand{\basepath}{graph-embedding/orkut}
  \newcommand{\orkutTot}[1]{Orkut_lambda-10.0_d-#1_gaussian-1.000000e-04_maxIter-8000_earlyExag-2000_seed-0}
  \newcommand{\orkutSub}[1]{\orkutTot{2}_with-comm_id-#1}
  \newcommand{\orkutHist}{orkut_degrees-histogram}
  % 
  %
  % ~~~~~ ORKUT
  %
  % 
  % \blocksec{Figure: real-world graphs}{6cm}
  \begin{subfigure}{\linewidth}
    \centering
    \begin{subfigure}{.35\linewidth}
      \centering
      \includegraphics[width=\linewidth]{%
        {{\basepath/\orkutTot{3}}}}
    \end{subfigure}
    \hspace*{\fill}
    \begin{subfigure}{.26\linewidth}
      \centering
      \includegraphics[width=\linewidth]{%
        {{\basepath/\orkutTot{2}}}}
    \end{subfigure}
    \hspace*{\fill}
    \hspace*{\fill}
    \hspace*{\fill}
    \begin{subfigure}{.28\linewidth}
      \centering
      \input{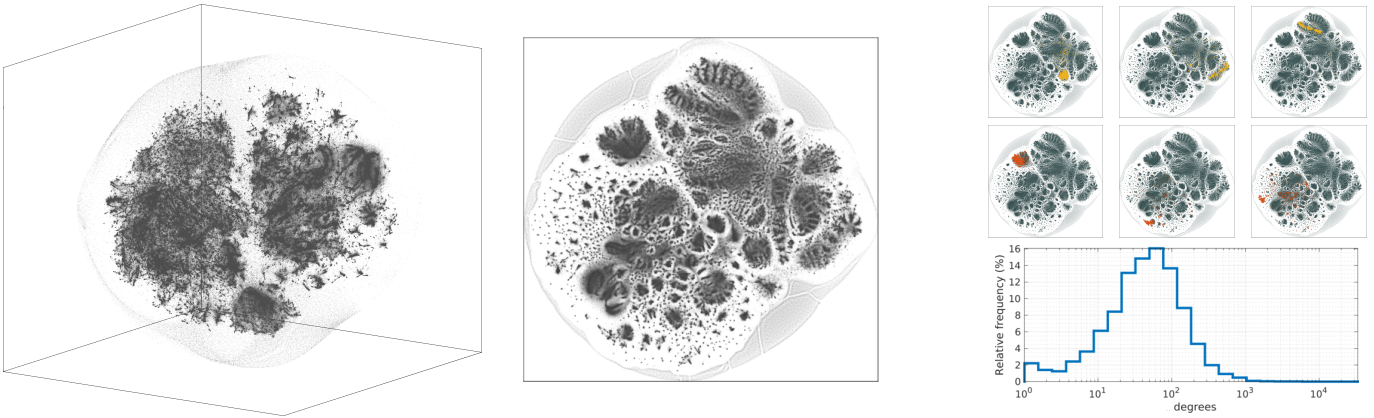}
    \end{subfigure}
  \end{subfigure}
  \caption[Orkut embedding]{\footnotesize %
    Spaceland (3D) and Flatland (2D) embedding of the friend social network
    \texttt{orkut} as of 2012~\cite{Yang2015}. The network has
    $\nNode = 3{,}072{,}441$ user nodes,
    $\nEdge = 237{,}442{,}607$ friendship links. The degree varies
    from $1$ (leaf nodes) to $33{,}313$. 
    The graph is not admissible to the conventional \tsne. 
    The embedding is by \sgtsne\  with $\lambda = 10$, taking $8{,}000$
    iterations, with initial coordinates drawn from a Gaussian random
    distribution.  The 2D embedding takes only
    50 minutes on the Xeon E5 (see \cref{tab:cpu-specs}).
    \textbf{Left\&Middle}: The vertex locations are structured,
    with entropy equal to $7.64$. The leaf nodes ($67{,}767$ of them) are 
    in the halo-like peripheral area. The rest can be
    roughly put into two hemispherical regions,
    which may likely correspond to the largest user populations in
    two geographical areas, one is in and around Brazil, the other
    is in and around India and some other countries in Asia.
    \textbf{Right}: The scatter diagrams in the thumbnails show six
    identified communities with labels {107, 325, 688} (top row) and
    {2196, 3208, 3405} (bottom row).  The top ones reside in
    one hemispherical region; the bottom ones, in the other. This
    pattern indicates the impact of geographical regions, cultures and
    societies on social network  structures.
  }
  \label{fig:orkut-2Dembedding}
\end{figure*}

%%% Local Variables:
%%% mode: latex
%%% TeX-master: "../HPEC2019-TSNEPI-main"
%%% End:

% 
% ================================================

We make a couple of remarks on the relationship between $u$ and $k$.
Commonly, the heuristic rule $k= \lfloor 3u \rfloor$ is
used. Veritably, the condition $0<u<k$ is necessary for the solutions
to \cref{eq:sigma-equalization} to exist at all vertices.  With a
larger perplexity $u$, one would increase $k$ proportionally to get a
better embedding in the sense of \cref{eq:distribution-matching-kl},
at the expense of a denser graph, larger memory requirement and longer
execution time.

To unleash the greater potential with the essential \tsne{}, we must
unravel two elaborate restraining knots in the standard form. One is
the coupling between distance and stochastics by
(\ref{eq:d-to-p}). The other is the pairing between vertex degree and
perplexity by \cref{eq:sigma-equalization}.

%%% Local Variables:
%%% mode: latex
%%% TeX-master: "../HPEC2019-TSNEPI-main"
%%% End:

%  LocalWords:  Kullback Leibler kNN Hadamard

% SG-as-interface.tex
%
% drafted by Xiaobai Sun on May 19, 2019
% editing by Alexandros (May 24, 2019)
%

\subsection{Advantages of admitting stochastic graphs}
\label{subsec:SG-abstraction}

Stochastic graph data arise in numerous and various situations.
Embedding of stochastic graphs will benefit network analysis and
network-based data analysis in general.

The stochastic kNN graphs obtained by \cref{eq:d-to-p} and
\cref{eq:sigma-equalization} are only a subset of stochastic graphs.
While they play a key role for \tsne\ being seen and used as a tool
for nonlinear dimensionality reduction and data visualization on the
one hand, they obscure the essential principle of \tsne{} in several
ways on the other hand.  
\begin{inparaenum}[(i)]
\item Real-world graph data stem from various 
  sources~\cite{kunegis2013konect, Kwak10www, icwsm10cha, auer2007dbpedia,
    Yang2015, yu2008high, redner2004citation, albert1999internet,
    zachary1977information, girvan2002community, newman2006finding,
    watts1998collective}.  Often, the native and interpretable descriptors
  of the entity vertices are not readily in a metric space.  The
  descriptors may contain both numerical and categorical data attributes.
  % For example, on a network for population study, the attribute
  % record for each individual may contain numerical values of age,
  % height, blood pressure and non-numerical information on
  % occupation, religion denomination and hobbies. The number of
  % attribute components may be very large.
  %
  For such data, \tsne\ may be preceded by a high-dimensional metric-space
  embedding, using machine learning tools (e.g., neural networks) and
  techniques~\cite{Mikolov2013,Pennington2014}, as in parametrized
  \tsne~\cite{Maaten2009}, at a very high computational cost.
  %
  % A more serious concern is that unsupervised feature
  % transformation is fundamentally driven by matching a reconstructed
  % graph from the transformed feature vectors to the data graph from
  %  the native features. The data graph is often the entity-attribute
  % bipartite.
%% 
\item In the case where the feature descriptors are in a metric space,
  constructing the \knn\ graph is computationally expensive.
  Randomized methods are used to reduce the asymptotic complexity to
  $O(n \log n)$~\cite{Muja2014,Muja2009}.  However, such methods do not leverage
  domain-specific spaces of random variables~\cite{Consortium2011}.
\item The \knn\ graph and the outcome of \tsne\ vary with the choice of
  vector space, distance function, and \knn\ search methods.
\item The stochastic recasting of a \knn\ graph is exclusively confined in
  conventional \tsne\ to the conditional Gaussian distribution, not
  inclusive of other probabilistic models.
\end{inparaenum}
The above limitations are common to many \tsne\
variants~\cite{Maaten2008,Amir2013,VanDerMaaten2014,Abdelmoula2016,%
  Pezzotti2016,Pezzotti2017,Unen2017,Linderman2019} and other
multi-dimensional scaling (MDS)
algorithms~\cite{Mead1992,fua1999hierarchical,huang2005multidimensional}.
There exist nonlinear dimension reduction methods that attempt to
circumvent the \knn\ search with certain sampling
strategies~\cite{mcinnes2018umap}.

Stochastic graphs are found or obtained in a multititude of other
ways.  We give two examples.
First, for high-security information networks, the vertex content features
are not publicly accessible and their links are stochastically
encrypted~\cite{Su2015}.
% in order to protect individual features and interaction
% links or messages from being exposed or recovered in network analysis.
%
Second, network data are often available only with binary-valued links.
Nonetheless, the stochastic weights of the links may be recovered to
significant extent from the local neighborhood topologies and the global
network topology.  We use in this paper two such stochastic graphs from
a recent study~\cite{liu2019density}: one is of a social network, the
other of an e-commerce network.
In addition, real-world graphs are typically irregular, with high-degree
hub nodes and 1-degree leaf nodes.  See for example the degree distribution
of the social network in \cref{fig:orkut-2Dembedding}.

In short, with stochastic graphs, we effectively create a clear
abstraction of distinguishing graph embedding from graph construction.
We also establish an interface between the two key processes,
disentangling them for integrating the best in each.  We explore the
deep potential of the \tsne{} essence.

%%% Local Variables:
%%% mode: latex
%%% TeX-master: "../HPEC2019-TSNEPI-main"
%%% End:

%  LocalWords:  parametrized MDS

% ==================================== figure E18-MBCs
% 
% fig-E18-MBCs-embedding.tex
%
% drafted by xiaobai on May 24, 2019

% ========== FIGURE: RNA SEQUENCE RESULTS
% 
\begin{figure*}
  \centering
  \newcommand{\basepath}{e18}
  \newcommand{\embed}[1]{e18mice-zheng-pca-50-exact_1306127_u-30_nn-90_d-#1_uniform-1_maxIter-8000_earlyExag-2000_seed-0}
  \newcommand{\recallplot}{e18mice-zheng-pca-50-exact_1306127_recall-histogram}
  \begin{subfigure}{.3\linewidth}
    \centering
    \includegraphics[width=\linewidth]{%
      \basepath/\embed{3}}
  \end{subfigure}
  \hspace*{\fill}
  \begin{subfigure}{.22\linewidth}
    \centering
    \includegraphics[width=\linewidth]{%
      \basepath/\embed{2}}
  \end{subfigure}
  \hspace*{\fill}
  \begin{subfigure}{.28\linewidth}
    \centering
    \includegraphics[width=\linewidth]{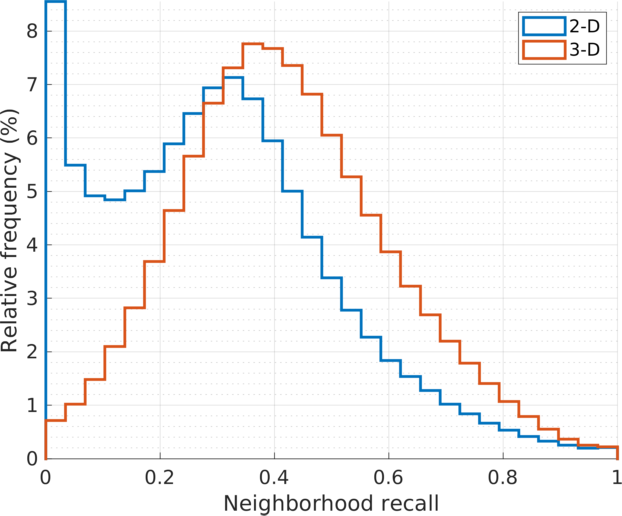}
  \end{subfigure}
  \caption[Embedding of E18 mouse brain cells.]{\footnotesize
    Spaceland(3D) and Flatland(2D) embedding of the kNN graph ($k=90$)
    associated with $\nNode = 1{,}306{,}127$ RNA sequences of E18
    mouse brain cells~\cite{BrainCells2017}.  The $k$-neighbors
    ($k=90$) are located in the space of the $50$ principle components
    of the top $1{,}000$ variable genes~\cite{Zheng2017}.
    \textbf{Left\&Middle}: The
    embeddings are obtained by \tsnepi{} with perplexity $u=30$;
    taking $8{,}000$ iterations, including $2{,}000$ early exaggeration
    iterations with $\alpha = 12$. The initial coordinates are drawn from
    a random uniform distribution. 
    Two subpopulations are color annotated:  
    GABAergic subtype (Sncg, Slc17a8) in blue and VLMC subtype (Spp1,
    Col15a1) in red. They are clearly separated from the rest in 3D
    embedding, see steerable 3D rendering in Supplementary Material. 
    \textbf{Right}: Histograms of the stochastic $k$-neighbors recall
    values at all vertices,
    $ {\rm recall}(i) = \sum_j \mathbf{P}_{j|i} \, b_{ij}$, where
    $\mathbf{B}_{k} = [b_{ij}]$ is the adjacency matrix of the $k$NN
    graph of the embedding points $\{\mathbf{y}_{i} \}$, in Euclidean
    distance. The stochastic weights $p_{j|i}$ are common to both 2D
    and 3D embedding.  The histogram for the 3D embedding (in red)
    shows relatively higher recall scores in larger cell
    population. The quantitative comparison is consistent with that by
    visual inspection.}
  \label{fig:embeddings-E18-MBC-RNA-sequences}
\end{figure*}

%%% Local Variables:
%%% mode: latex
%%% TeX-master: "../HPEC2019-TSNEPI-main"
%%% End:

% 
% ================================================

% SG-t-SNE-description.tex
%
% drafted by Xiaobai Sun on May 19, 2-19
%
%
% Revised by Dimitris F. on May 24, 2019
% editing by Alexandros (May 24, 2019)

\subsection{Introduction of SG-t-SNE}
\label{subsec:SG-t-SNE-introduction}

Our algorithm \sgtsne\ follows the essential principle of \tsne\ and
removes the barriers in its conventional version. \sgtsne{} admits any
stochastic graph $\graph = (\vertices, \edges, \weightMatCond)$,
including stochastic \knn\ graphs.  \sgtsne\ is capable of not only
adapting to the sparse pattern in the input, but also exploiting the
pattern for better distribution matching and vertex embedding with a
non-linear re-scaling mechanism.

Specifically, we generate from $\weightMatCond$ a parameterized family
of stochastic matrices, $\weightMatCond(\lambda)$, via nonlinear
rescaling.  The role of the rescaling mechanism is to explore and
exploit the potential in distribution matching
\cref{eq:distribution-matching-kl} from the $\weightMat$ side, given
the fixed t-distribution (a Cauchy distribution) on the
$\weightMatEmb$ side.  Unlike the perplexity $u$ in the conventional
t-SNE, the parameter $\lambda$ for SG-t-SNE exploits the sparsity
without imposing any constraint.  In the special case of a weighted
\knn\ graph, $\lambda$ is untangled from $k$.

We describe the mechanism and effect of nonlinear rescaling with a single
parameter $\lambda$.
Let $\phi(\cdot)$ be a rescaling kernel function that is monotonically
increasing over $\mathbb{R}_{+}$, with $\phi(0) \geq 0$.
%
% \xsnote{\em For later longer documentation: the entropy function does
%   not meet this condition.}
%
For any $\lambda > 0$, \sgtsne\ generates $\weightMatCond(\lambda)$ as
follows.  Let $\mathbf{A}$ be the binary-valued adjacency matrix
(i.e., the sparsity mask) of graph $\graph$, which is invariant to any
change in $\phi$ or $\lambda$.  We determine the rescaling exponent
$\gamma_i$ at each vertex $i$ by the equation
\begin{equation}
  \label{eqn:rescaling-exponens} 
  \sum_{j} a_{ij} \, \phi\left( p_{j|i}^{\gamma_i} \right) = \lambda , 
  % \quad \forall i \in \{1, \ldots, n\} ,
\end{equation}
and then reshape and rescale the conditional probability
\begin{equation}
  \label{eqn:reshaping-rescaling} 
  p_{j|i}( \lambda ) =
  \frac{ a_{ij} \, \phi\left( p_{j|i}^{\gamma_i} \right) }{\lambda}.
\end{equation}
Unlike \cref{eq:sigma-equalization}, the rescaling solutions exist
unconditionally. 

The rescaling mechanism introduces an additional degree of freedom to
exploit the sparsity in an arbitrary form.  In particular, we use the
identity function $\phi(x)=x$ as the rescaling kernel for the
experiments presented in this paper.  At $\lambda = 1$ we get
$\gamma_i=1$ and $\weightMatCond(1) = \weightMatCond $, the input
matrix. At an integer value $\lambda = k > 1$, if
$\mbox{degree}(\ell) = k$, then $\gamma_{\ell} = 0$ and
$p_{j|\ell} = 1/k$.  In the special case of a regular graph with
degree $k$, if we set $\lambda =k$, then $\gamma_i=0$ at all vertices,
i.e., $\weightMatCond(k) = \mathbf{A}/k$.
In general, at $\lambda \neq 1$, the ratio between every nonzero
element to the peak element $ \left( p_{j|i}/\max_{l}p_{l|i} \right)$
is changed to $ \left( p_{j|i}/ \max_{l}p_{l|i} \right)^{\gamma_i}
$. The weight distribution for each vertex is re-shaped, according to
the re-scaling equation (\ref{eqn:rescaling-exponens}), in autonomous
adaptation to the local sparsity.

% becomes flatter with $\lambda >1$ and more elevated at the peak
% point with $\lambda <1$.
%
% As $\lambda \to \infty$, $\weightMatCond(\lambda)$ approaches the
% degree-normalized adjacency matrix.  As $\lambda \to 0$,
% $\weightMatCond(\lambda)$ approaches the adjacency matrix of the
% neighbor graph (which need not be regular).

In implementation, only two changes are made from the conventional
version.  One is at the interface, which takes either a distance-based
\knn\ graph as in the conventional \tsne\ or a stochastic matrix
directly. The other is the change from the perplexity equation
\cref{eq:sigma-equalization} to the rescaling equation
\cref{eqn:reshaping-rescaling}.

\begin{figure}
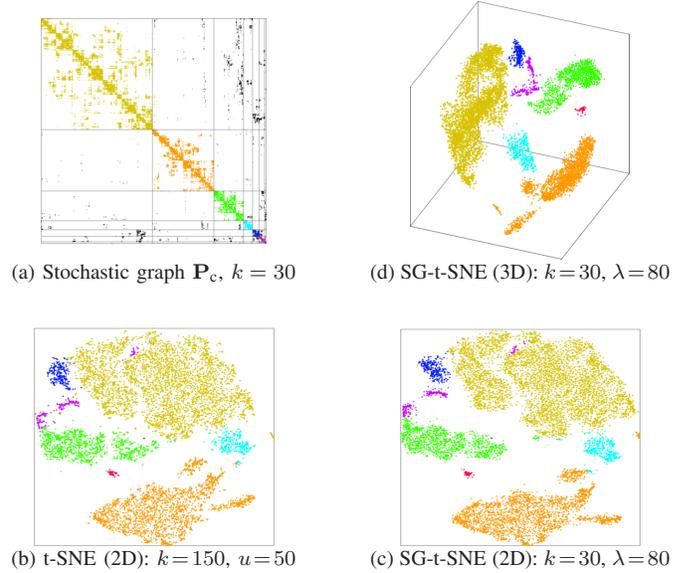

  \centering
  % --- COMMANDS
  % 
  \newcommand{\basepath}{pbmc}
  \newcommand{\embedLambda}[1]
  {pbmc_lambda-80.0_nn-30_d-#1_uniform-1_maxIter-1000_earlyExag-250_seed-0}
  \newcommand{\embedU}[1]
  {pbmc_u-50_nn-150_d-#1_uniform-1_maxIter-1000_earlyExag-250_seed-0}
  % 
  % 
  % --- FIGURES
  % 
  \begin{subfigure}{.45\linewidth}
    \centering
    \includegraphics[width=0.75\linewidth]{%
      {{\basepath/pbmc-gk-binary}}}
  \end{subfigure}
  \hspace*{\fill}
  \begin{subfigure}{.45\linewidth}
    \centering
    \includegraphics[width=.73\linewidth]{%
      {{\basepath/\embedLambda{3}}}}
  \end{subfigure}
  \\[0.1em]
  % 
  % 
  % --- CAPTIONS
  % 
  \begin{subfigure}{.45\linewidth}
    \centering
    \caption*{\footnotesize (a) Stochastic graph $\mathbf{P}_\text{c}$, $k=30$}
  \end{subfigure}
  \hspace*{\fill}
  \begin{subfigure}{.45\linewidth}
  \caption*{{\footnotesize (d) \sgtsne{} (3D): $k\!=\! 30$, $\lambda\!=\! 80$}}
  \end{subfigure}
  \\[1em]
  % 
  % ===== SECOND ROW
  % 
  % --- FIGURES
  % 
  \begin{subfigure}{.45\linewidth}
    \centering
    \scalebox{1}[-0.87]{
      \includegraphics[width=.8\linewidth]{%
        {{\basepath/\embedU{2}}}}
    }
  \end{subfigure}
  \hspace*{\fill}
  \begin{subfigure}{.45\linewidth}
    \centering
    \scalebox{1}[-0.8]{
      \includegraphics[width=.8\linewidth]{%
        {{\basepath/\embedLambda{2}}}}
    }
  \end{subfigure}
  \\[0.1em]
  % 
  % 
  % --- CAPTIONS
  % 
  \begin{subfigure}{.45\linewidth}
    \centering
    \caption*{\footnotesize (b) \tsne{} (2D): $k\!=\! 150$, $u\!=\! 50$}
  \end{subfigure}
  \hspace*{\fill}
  \begin{subfigure}{.45\linewidth}
    \centering
    \caption*{\footnotesize (c) \sgtsne{} (2D): $k\!=\! 30$, $\lambda\!=\! 80$}
  \end{subfigure}
  \caption[PBMC embedding results]{\footnotesize %
    Multiple views of subpopulations of $n \!=\! \num{8 381}$
    peripheral blood mononuclear cells
    (\texttt{PBMC-8k})~\cite{Zheng2017}.
    The
    subpopulations (color-coded) were found by the SD-DP
    classification analysis~\cite{Floros2018}.
    % with threshold $\tau=0.3$, {\bf which k value ??}
    % 
    \textbf{(a)}~The stochastic $k$NN graph $\mathbf{P}_\text{c}$ ($k=30$),
    with rows/columns
    permuted so that cells in the same subpopulation are placed
    together. Each pixel corresponds to a $16\times 16$ matrix block.
    % The column are color coded.
    % {\bf off-diagonal elements shall not be color coded. --xs}
    %
    \textbf{(b)\&(c)} It is compelling that the  embedding by
    \sgtsne{} on a sparse graph is comparable to that by \tsne{}
    on a graph $5\times$ as dense.
    \textbf{(d)}~Rapid Spaceland (3D) embedding is enabled by \tsnepi{},
    steerable view available in Supplementary Material. 
    All three embeddings take $\num{1000}$ iterations, including $250$
    early exaggeration iterations with $\alpha = 12$, the initial
    embedding coordinates are drawn from uniformly random
    distribution.} 
  \label{fig:pbmc}
\end{figure}

We show in \cref{fig:amazon,fig:orkut-2Dembedding} the embeddings of
two real-world networks, enabled by \sgtsne{}. Each exhibits a
distinctive cluster structure.
We show in \cref{fig:pbmc} that even with kNN graphs at input, the
embedding by \tsne{} at $k=150$, $u=50$ is matched or outperformed by
that with \sgtsne{} at $k=90$, $\lambda=10$, on a sparser matrix and
at a lower computation cost. We recommend multiple views at various
values of $\lambda$.

% The following is useful: 
% We comment in passing that the trajectory in embedding coordinates and the
% morphism between embedding patterns give additional quantitative information
% to the graph under study and may lead to better interpretation or
% hypotheses.

%%% Local Variables:
%%% mode: latex
%%% TeX-master: "../HPEC2019-TSNEPI-main"
%%% End:

%  LocalWords:  rescaling rescale

\section{Rapid spaceland embedding}
\label{sec:spaceland-embedding}

% spaceland-challenges.tex

The practical use of \tsne\ is largely confined to Flatland (2D) or
Lineland (1D) embedding, despite the prototype implementation by van
der Maaten%
\footnote{\url{https://lvdmaaten.github.io/tsne}} %
supporting embedding in several dimensions.
We advocate Spaceland (3D or higher) embedding.  Among abundant
evidence, we illustrate in
\cref{%
fig:embedding-with-synthetic-data,%
fig:orkut-2Dembedding,%
fig:embeddings-E18-MBC-RNA-sequences}, %
the extended capacity to capture and reveal multi-fold connections
between (overlapping) subpopulations. With Spaceland embedding, we
endorse ``the enlargement of the imagination''~\cite{flatland}.

There were multiple obstacles to Spaceland embedding.
% 
% There are $d n$ coordinate parameters to be determined in the embedding
% $\{\mathbf{y}_i\}$.
The journey of \tsne\ is marked by continuous efforts and progress in
reducing the arithmetic complexity and execution time of searching for
an optimal embedding.
We present \tsnepi, a software renovation, a new milestone in
performance.  It enables rapid Spaceland embedding of large sparse
graphs on modern desktop or laptop computers, which are available and
affordable to researchers by and large.

The embedding coordinates,, i.e., $d\, n$ parameters, are determined
by \tsne{} via applying the gradient descent method to the
minimization problem \cref{eq:distribution-matching-kl}.  The
computation per iteration step is dominated by the calculation of the
gradient, which is reformulated in two terms by van der
Maaten~\cite{VanDerMaaten2014},
\begin{equation}
  \label{eq:gradient-in-2terms}
  \frac{ \partial( \opKL ) }{ \partial \mathbf{y}_i }
  =
  \frac{4}{Z}
  % \Bigg(
  \underbrace{
    \sum_{i\neq j} p_{ij} q_{ij} (\mathbf{y}_i - \mathbf{y}_j)
  }_{\Fattr}
  -
  \frac{4}{Z}
  \underbrace{
    \sum_{i\neq j} q^2_{ij} (\mathbf{y}_i - \mathbf{y}_j),
  }_{\Frep}
  % \Bigg)
\end{equation}
where $Z=\sum_{ij}q_{ij}$.  The terms are named by analogy to the
attractive and repulsive forces in molecular dynamics simulations.
The attractive term can be cast as the sum of multiple matrix-vector
products with a sparse matrix $\mat[PQ] = [p_{ij}q_{ij}]$ and $d$
coordinate vectors, one along each embedding dimension.  Similarly, the
repulsive term can be cast as the sum of multiple matrix-vector products
with a dense matrix $\mat[QQ] = [q_{ij}q_{ij}]$ and $d$ coordinate vectors.
The calculation of the attractive term takes $O(d m)$ arithmetic
operations, where $m=|E|$ is the number of edges.  A naive calculation of
the repulsive term takes $O(dn^2)$ arithmetic operations. This quadratic
complexity would prohibit the use of \tsne\ with large graphs.

%%% Local Variables:
%%% mode: latex
%%% TeX-master: "../HPEC2019-TSNEPI-main"
%%% End:

% QQ-term-approximation.tex

\subsection{Fast approximations to the repulsive term}
\label{subsec:BH-FI}

Two existing \tsne\ variants use approximate algorithms to calculate
the repulsive term.  They reduce the quadratic complexity to
$O(c^d(\epsilon^{-1}) \, n \log(n))$, where $c(\epsilon^{-1}) > 2$
increases reciprocally with $\epsilon$, a prescribed approximation
error tolerance.  The specific value of $c$ in the prefactor $c^d$
depends also on the choice of approximation algorithm and its
parameters.
The first approximation algorithm, by van der Maaten, adopts the tree code
based on the Barnes-Hut (BH) algorithm~\cite{Barnes1986}.  We refer to
this version as \tsnebh.
An alternative approximation algorithm, named \tsnefi, emerged last
year~\cite{Linderman2019}.

We describe the fundamental concepts and distinctive approaches behind
the two approximation algorithms, instead of reviewing each in detail,
The matrix-vector multiplication with $\mat[QQ]$ is a convolution with the
Cauchy kernel on non-equispaced, scattered data points
$\{\mathbf{y}_i\}_{i=1}^n$.  We refer to such a convolution as ``nuConv''.
A convolution may have broad or narrow (windowed) support.
Narrow-support convolution has complexity $O(w n)$, where $w$ is the
bandwidth or support window size; the corresponding matrix is sparse and
banded.
The nuConv with $\mat[QQ]$ is of broad support and dense.
Fortunately, the matrix structure can be exploited.

There are two renowned and distinctive families of algorithms for fast
nuConv of broad support.  One family is based on the fast multipole
method (FMM), with asymptotic arithmetic complexity
$O(n)$~\cite{Greengard1987,Sun2001,zhang2016rec}.  FMM splits a
convolution of broad support into $O(\log(n))$ convolutions of narrow
supports at multiple spatial levels. Its extended family includes the
Barnes-Hut algorithm, with complexity $O(n\log(n))$~\cite{Barnes1986}.
The other family is based on non-uniform fast Fourier transforms
(nuFFTs), with arithmetic complexity
$O(n\log(n))$~\cite{wang2014generalized,Zhang2012}.  \tsnefi\ takes
the latter approach.

\tsnefi\ is available for 1D/2D embedding only. By a careful comparison,
its arithmetic complexity is lower than \tsnebh\ at the same accuracy
level.  Nonetheless, the execution time for 2D embedding with \tsnefi\ (or
\tsnebh) is now dominated by the computation of the attractive term; see
\cref{fig:exec-time-tsnepi-tsnefi}.  Furthermore, both \tsnefi\ and
\tsnebh\ face a steep rise in execution time for 3D embedding.  \tsnefi\
also suffers from exponential growth in memory demand due to zero
padding for explicit conversion to periodic convolution.

% ==================== FIGURE: AMAZON
% 
% fig-pbmc.tex
%

% Drafted by Dimitris F. on May 20, 2019
%
% caption revised by Xioabai on May 25, 2019
%
% caption/subcaption revised by Xioabai on June 3, 2019 :
% ( after a skype meeting with Dimitris and Nikos to sort out
% a couple of issues: re-clustering, re-coloring, exploring closer
% neighborhood )
%
% caption/subcaption revised by Xioabai on June 4, 2019 :
%

% 
% 
%%% Local Variables:
%%% mode: latex
%%% TeX-master: "../HPEC2019-TSNEPI-main"
%%% End:

%%% Local Variables:
%%% mode: latex
%%% TeX-master: "../HPEC2019-TSNEPI-main"
%%% End:

% PI-t-SNE-introduction.tex
%
% drafted by Xiaobai Sun on May 19, 2-19
% editing by Alexandros (May 24, 2019)
% revised by Xiaobai on June 2. 2019
% 

\subsection{Introduction of \tsnepi}
\label{subsec:tSNE-PI-introduction}

With \tsnepi, we advance the entire gradient calculation of
\cref{eq:gradient-in-2terms} and enable Spaceland~(3D) embedding in
shorter time than Flatland~(2D) embedding with \tsnefi\ or \tsnebh.
We overcome challenges at multiple algorithmic stages with innovative
approaches. We utilize the matrix structure and the memory
architecture in concert.

We not only expedite further the repulsive term computation with dense
matrix $\mat[QQ]$, but also accelerate the attractive term computation
with sparse matrix $\mat[PQ]$.  Furthermore, we make swift
translocation of $\{\mathbf{y}_i\}$ between the two terms.

% In the rest of this section, we briefly describe the key
% building blocks of \tsnepi.

%%% Local Variables:
%%% mode: latex
%%% TeX-master: "../HPEC2019-TSNEPI-main"
%%% End:

% PI-t-SNE-qq.tex
%
% drafted by Xiaobai Sun on May 19, 2-19
% editing by Alexandros (May 24, 2019)
% revised by Xiaobai on June 3, 2019
% 

\subsubsection*{Fast computation with $\mat[QQ]$}
\label{sec:fast-QQ}

We take the spectral approach, similar to FIt-SNE.  Set first an
equispaced grid $G$ in the spatial embedding domain. The nuConv with
$\mat[QQ]$ is then factored into three consecutive convolutional
operations which we code-name as S2G, G2G and, G2S.
The S2G and G2S convolutions translate $\{\vec[y]_i\}$ to their
neighboring grid points, and vice versa.  G2S is instrumented as a
local interpolation and S2G is its transpose. Both are of narrow
support, with arithmetic complexity $O(w^d n)$, where $w$ is the
interpolation window size per side. Factor G2G is a non-periodic
convolution across the grid $G$.

We make algorithmic innovations for each of the factor convolutions.
Instead of explicit embedding G2G into a circulant one as in FIt-SNE,
we use an implicit approach without augmenting the grid size and its
memory usage by a factor of $2^d$, while maintaining the same
arithmetic complexity. This memory saving allows a laptop or desktop
computer to admit larger graphs.
The factors S2G and G2S prolong the execution time due to irregular
data access, a typical challenge in sparse matrix calculation.  We
introduce a dual grid $G_{\rm dual}$ for gridding the scattered data
points $\{\mathbf{y}_i\}$.  Prior to S2G, the scattered points are
binned into the dual-grid cells. Then, S2G translates the gridded data
on $G_{\rm dual}$ to grid $G$, on which $G2G$ takes place. This
approach significantly improves the data locality and parallelism
scope in S2G, and shortens the S2G time, accounting for the gridding
overhead.  The same holds true for G2S.
We omit implementation details.

%%% Local Variables:
%%% mode: latex
%%% TeX-master: "../HPEC2019-TSNEPI-main"
%%% End:

% fig-perf-bar-charts.tex
%
% Drafted by Dimitris F. on May 22, 2019
%
\begin{figure}
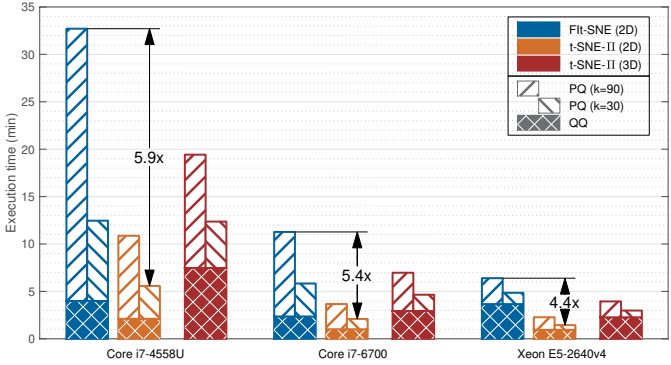

  \centering
  %
  % --- 5 modules
  %
  %    \includegraphics[width=\figbig]{%
  %    {{tsnepi-performance/tsne-perf-comp_e18mice-zheng-pca-50-exact_1306127}}}
  %
  % --- 2 modules
  %
  % SCRIPT: code/visualize_tsne_performance_comparison.m
  % 
  \includegraphics[width=\figbig]{%
  {{tsnepi-performance/tsne-perf-comp-two-modules_e18mice-zheng-pca-50-exact_1306127}}}
  \caption[Comparison in execution time]{\footnotesize 
    Multiple comparisons in execution time for embedding of $k$NN
    graphs, $k\in \{30,90\} $, with $\num{1 306 127}$ nodes as 
    single-cell RNA sequences of E18 mouse brain cells~\cite{BrainCells2017}. Each
    embedding takes $\num{1000}$ iterations and maintains below the same
    tolerance ($10^{-6}$) on approximation errors.
    \begin{inparaenum}[(a)] 
     \item
      In total time, with the same sparsity parameter $k$, 3D embeddings
      by \tsnepi{} take less time to finish $1.5\times$ operations in
      comparison to 2D embeddings by \tsne{}, on each of
      the three multi-core computers used.
    \item
      In memory usage, \tsnepi{} does not augment grid size
      and thereby admits larger graphs. 
    \item
      The computation with sparse matrix $\mathbf{PQ}$
      dominates the \tsne{} time on machines with lower bandwidth,
      see \cref{tab:arch-specifications}.
      \sgtsne{} reduces the PQ time further ($5\times$
      on Core i7-6700) by embedding a sparse graph ($k=30$)
      with results comparable to that by \tsne{} on the graph $3\times$ as dense.  
    \end{inparaenum}
    Additional notes: the \tsnepi{} code is to be optimized; a 
    version utilizing graphics card, if any, is in development.  }
  \label{fig:exec-time-tsnepi-tsnefi}
  % \vspace*{-1em}
\end{figure}
%
%%% Local Variables:
%%% mode: latex
%%% TeX-master: "../HPEC2019-TSNEPI-main"
%%% End:

% PI-t-SNE-pq.tex
%
% drafted by Xiaobai Sun on May 19, 2-19
% editing by Alexandros (May 24, 2019)
% revised by Xiaobai Sun on June 4, 2019
% 

\subsubsection*{Fast computation with $\mat[PQ]$}
\label{sec:fast-PQ}

Irregular sparse pattern of $\mat[PQ]$ invokes irregular memory
accesses which incur long latency on modern computers with
hierarchical memories~\cite{Mellor-Crummey2001}.  The execution time of \tsnefi\ is
evidently dominated by the computation with $\mat[PQ]$ at $k=90$ and
$u=30$; see \cref{fig:exec-time-tsnepi-tsnefi}.
The CPU is frequently in data-hungry state, the computation speed is
limited by the memory bandwidth.

We mitigate the problem with a preprocessing step, based on the fact
that matrix $\mat[PQ]$ inherits the fixed sparsity of matrix
$\mat[P]$, despite the dense structure and dynamic change of
$\mathbf{Q}$.
% 
% The arithmetic complexity for the computation with $\mat[PQ ]$ is
% $O(d\, m)$, $m = |\edges|$.  When $\mathbf{P}$ is a \knn\ graph,
% $m=k n$.
%
We permute $\mathbf{P}$ such that similar rows and columns are
clustered and placed together. The permuted matrix becomes
block-sparse with denser blocks (BSDB), not necessarily block
diagonal.
%
% When the inter-connection between the clusters is weak, the matrix
% is nearly block diagonal.
%
A denser block implies higher data locality in memory reads and
writes.  We employ several efficient ways for clustering rows and
columns (symmetrically). For a \knn\ graph at input, in particular,
\tsnepi{} has options among PCA-based cluster analysis and the SD-DP
classification analysis~\cite{Floros2018}.
%
% For other types of sparse graphs, \tsnepi{} employs the new
% algorithm in \cite{tiancheng}.
%
The permutation overhead is amortized across all iterations. It is
well paid off; see \cref{fig:exec-time-tsnepi-tsnefi}.

In implementation detail, we adopt the storage format and sparse
matrix computation routines provided in the Compressed Sparse Blocks
(CSB) library~\cite{Buluc2009, Blumofe1996}.

%%% Local Variables:
%%% mode: latex
%%% TeX-master: "../HPEC2019-TSNEPI-main"
%%% End:

% ---------- TABLE: CPU specs

% tab-CPU-specs.tex
% 
%
% drafted by xiaobai on May 24, 2019

% ==================== TABLE: CPU SPECS

\begin{table}
  \centering
  \caption[CPU specifications]{\footnotesize %
    Archtecture specification of three multi-core computers
    used for the embedding experiments.
    Memory bandwidth is measured with the parallel \texttt{STREAM}
    \texttt{copy} benchmark~\protect{\cite{stream}}.}
  \label{tab:cpu-specs}
  \begin{adjustbox}{max width=\linewidth}
    % table-cpu-specs.tex
%
% created and revised by Alexandros
%
%
\newcommand{\hcolsep}{\hspace{3pt}}
\begin{tabularwithnotes}{%
    @{}
    l                   @{\hcolsep} % CPU
    S[table-format=1.2] @{\hcolsep} % clock
    S[table-format=2]   @{\hcolsep} % # cores
    S[table-format=2]   @{\hcolsep} % L1
    S[table-format=3]   @{\hcolsep} % L2 
    S[table-format=2]   @{\hcolsep} % L3
    % S[table-format=3]   @{\hcolsep} % L4
    S[table-format=3]   @{\hcolsep} % RAM
    S[table-format=2.1]             % STREAM B/W (parallel)
    % d{1.2} @{\hcolsep}              % clock
    % d{2.0} @{\hcolsep}              % # cores
    % d{2.0} @{\hcolsep}              % L1
    % d{3.0} @{\hcolsep}              % L2
    % d{2.0} @{\hcolsep}              % L3
    % d{3.0} @{\hcolsep}              % L4
    % d{3.0} @{\hcolsep}              % RAM
    % d{2.1}                          % STREAM B/W (parallel)
    @{}
    % @{\hspace*{-2.5pt}}
  }%
  %
  % === NOTES
  %
  {
    \tnote[$\ast$]{2 sockets (non-uniform memory access---NUMA)}
  }
  \toprule
  %
  % --- Columns
  %
    \multicolumn{1}{ @{} l @{\hcolsep} }{\textbf{CPU}}
  % & \multicolumn{1}{ @{} c @{\hcolsep} }{\textbf{Date}}
  & \multicolumn{1}{ @{} c @{\hcolsep} }{\textbf{Clock}}
  & \multicolumn{1}{ @{} c @{\hcolsep} }{\textbf{Cores}}
  & \multicolumn{1}{ @{} c @{\hcolsep} }{\textbf{L1}}
  & \multicolumn{1}{ @{} c @{\hcolsep} }{\textbf{L2}}
  & \multicolumn{1}{ @{} c @{\hcolsep} }{\textbf{L3}}
  % & \multicolumn{1}{ @{} c @{\hcolsep} }{\textbf{L4}}
  % & \multicolumn{1}{ @{} c @{\hcolsep} }{\textbf{CL}}
  % & \multicolumn{1}{ @{} c @{\hcolsep} }{\textbf{ISE}}
  & \multicolumn{1}{ @{} c @{\hcolsep} }{\textbf{RAM}}
  % & \multicolumn{1}{ @{} c @{\hcolsep} }{\textbf{BW}} % (max)
  & \multicolumn{1}{ @{} c @{}      }{\textbf{BW}} % (stream)
  %
  % --- Units
  %
  \\
    \multicolumn{1}{ @{} c @{\hcolsep} }{}
  % & \multicolumn{1}{ @{} c @{\hcolsep} }{}
  & \multicolumn{1}{ @{} c @{\hcolsep} }{\tiny (\si{\giga\hertz})}
  & \multicolumn{1}{ @{} c @{\hcolsep} }{}
  & \multicolumn{1}{ @{} c @{\hcolsep} }{\tiny (\si{\kibi\byte})}
  & \multicolumn{1}{ @{} c @{\hcolsep} }{\tiny (\si{\kibi\byte})}
  & \multicolumn{1}{ @{} c @{\hcolsep} }{\tiny (\si{\mebi\byte})}
  % & \multicolumn{1}{ @{} c @{\hcolsep} }{\tiny (\si{\mebi\byte})}
  % & \multicolumn{1}{ @{} c @{\hcolsep} }{(\si{\byte})}
  % & \multicolumn{1}{ @{} c @{\hcolsep} }{}
  & \multicolumn{1}{ @{} c @{\hcolsep} }{\tiny (\si{\gibi\byte})}
  % & \multicolumn{1}{ @{} c @{\hcolsep} }{(\gibi\byte\per\second)}
  & \multicolumn{1}{ @{} c @{}      }{\tiny (\si{\gibi\byte\per\second})}
  \\
  %
  % --- Sub-title (on columns)
  %
  & &
  & \multicolumn{2}{ @{} c @{\hcolsep} }{\footnotesize per-core}
  & \multicolumn{1}{ @{} c @{\hcolsep} }{\footnotesize shared}
  & &
  \\
  \midrule
  % 
  % ---------- MACBOOK LATE 2013 (DF)
  % 
  Intel Core i7-4558U
  & 2.80
  & 2 
  & 64
  & 256
  & 4
  % & \multicolumn{1}{c@{\hcolsep}}{\hspace*{-1ex}---}
  & 8
  & 10.5                        % direct-access STREAM
  \\
  % 
  % ---------- MACBOOK PRO LATE 2013 (AI)
  % 
  % Core i7-4960HQ
  % & 2.60
  % & 4
  % & 32
  % & 256
  % & 6
  % & 128
  % & 16
  % & 18.6
  % \\
  % 
  % ---------- BARITONE -- BASSOON -- CLARINET -- BONGO
  % 
  Core i7-6700
  % & Q3'15
  & 3.40
  & 4 
  & 32
  & 256
  & 8
  % & \multicolumn{1}{c@{\hcolsep}}{\hspace*{-1ex}---}
  % & 64
  % & AVX2
  & 32
  % & 34.1
  % & 0.34\text{(s)}              % indirect-access STREAM
  % & 1.02\text{(p)}              % indirect-access STREAM
  & 19.9                        % direct-access STREAM
  \\
  % 
  % ---------- DIGIPRO
  % 
  % \makecell[cl]{AMD Ryzen 1900X}
  % % & Q3'15
  % & 3.80
  % & 8 
  % & 96
  % & 512
  % & 16
  % & \multicolumn{1}{c@{\hcolsep}}{\hspace*{-1ex}---}
  % % & 64
  % % & AVX2
  % & 64
  % % & 34.1
  % % & 0.34\text{(s)}              % indirect-access STREAM
  % % & 1.02\text{(p)}              % indirect-access STREAM
  % & 31.6                         % direct-access STREAM
  % \\
  %
  % ---------- LINUX31-50
  %
  Xeon E5-2640v4\tmark[$\ast$]
  % & Q1'16
  & 2.40
  & 10
  & 32
  & 256
  & 25
  % & \multicolumn{1}{c@{\hcolsep}}{\hspace*{-1ex}---}
  % & 64
  % & AVX2
  & 256
  % & 68.3
  % & 0.62\text{(s)}              % indirect-access STREAM
  % & 2.46\text{(p)}              % indirect-access STREAM
  & 36.7                        % direct-access STREAM
  \\
  \bottomrule
\end{tabularwithnotes}%
%
%
%
%%% Local Variables:
%%% mode: latex
%%% TeX-master: "../HPEC2019-TSNEPI-main"
%%% End:

  \end{adjustbox}
  \label{tab:arch-specifications}
  %
  % \vspace*{-1em}
\end{table}
%
%%% Local Variables:
%%% mode: latex
%%% TeX-master: "../HPEC2019-TSNEPI-main"
%%% End:

% PI-t-SNE-description.tex
%
% drafted by Xiaobai Sun on May 19, 2-19
% editing by Alexandros (May 24, 2019)
% revised by Xiaobai Sun on June 4, 2019
% 

\subsubsection*{Fast data translocation}
\label{sec:fast-translocation}

The embedding coordinate data $\{ \mathbf{y}_i \}$ undergo dynamic
change during the process of iterative search and matching
\cref{eq:distribution-matching-kl}.  They are accessed in the
computation with both $\mat[QQ]$ and $\mat[PQ]$, but in different
orderings.
Matrix $\mat[P]$, and hence $\mat[PQ]$, is permuted, once for all, by
row/column clusters. The order $\{ \mathbf{y}_i \} $ is accessed
in the computation with $\mat[QQ]$, is determined by the dual-grid
cells for S2G and G2S.
At each iteration step, we translocate the data $\{ \mathbf{y}_i \}$,
back and forth between the two terms, in multiple stages.  In essence,
a point-to-point cross permutation is factored into layers of
block-to-block permutations.

We use $\Pi$ in the name of \tsnepi\ to signify the role of
permutations, factored permutation and the rational behind.
Permutations are unaccounted for in arithmetic complexity.  They play,
however, the key role of making the algorithm and its building blocks
attuned to the memory hierarchy. Without this harmonious
algorithm-architecture mapping, memory access latency typically
dominates the execution time in computation with large, sparse or
compressible matrices~\cite{patterson2014,yzelman2009cache}.
%

%%% Local Variables:
%%% mode: latex
%%% TeX-master: "../HPEC2019-TSNEPI-main"
%%% End:

% ==================== DISCUSSION
%
\section{Additional remarks}
\label{sec:discussion}
%
% discussion.tex
%

The embedding results obtained by \sgtsne{}, powered by \tsnepi{}, are
elucidating.
The embedding results with {\tt orkut} and {\tt Amazon} show clearly
distinctive cluster or communities patterns. Furthermore, we find
overlapping and non-overlapping regions between communities in {\tt orkut}.
The embedding of {\tt Amazon} may serve
as a map for market study and analysis, in assessing previous claimed
product clusters and making new hypotheses or predictions.
Even with $k$NN graphs, 3D embedding liberates us from 2D confinement
and takes us to a deeper space where we can see multiple views at
once, discover previously unseen connections and separations, such as
with the PBMCs data and the E18-MBCs data.
Additionally, we bring out the fascinating beauty of {\tt orkut} and
{\tt Amazon} in their respective embedding spaces.

%%% Local Variables:
%%% mode: latex
%%% TeX-master: "../HPEC2019-TSNEPI-main"
%%% End:

%%% Local Variables:
%%% mode: latex
%%% TeX-master: "../HPEC2019-TSNEPI-main"
%%% End:

\phantomsection
\label{sec:acknowledgments}
\addcontentsline{toc}{section}{Acknowledgments}
\section*{Acknowledgments}
%
% acknowledgments.tex
%
% Drafted by Dimitris F. on May 10, 2019

We thank Tiancheng Liu for providing the sparse and stochastic graphs
for the \texttt{orkut} and \texttt{Amazon} networks.
We thank Pantazis Vouzaksakis and Xenofon Theodoridis for their assistance
with the experiments.

%%% Local Variables:
%%% mode: latex
%%% TeX-master: "../HPEC2019-TSNEPI-main"
%%% End:

%%% =====================
%%% REFERENCES

% path to references file(s)

\newcommand{\pathRefs}{./HPEC2019-TSNEPI-main.bib}
% \nocite{*}                      % [REMOVE!]

% section/bookmark
\phantomsection
\label{sec:references}
\addcontentsline{toc}{section}{References}

%%%% ====================
%%%  bibliography

\ExplSyntaxOn
\str_case_x:nn {\g_format_str} {
  {plain} {                     % ========== PLAIN
    \ExplSyntaxOff
    \ifreview
    \bibliographystyle{apa}
    \else
    \bibliographystyle{abbrv}
    \fi
    \bibliography{\pathRefs}
    \ExplSyntaxOn }
  {hpec} {                      % ========== HPEC
    \ExplSyntaxOff
    \ifreview
    \bibliographystyle{apa}
    \else
    \bibliographystyle{IEEEtran}
    \fi
    \balance
    \bibliography{IEEEabrv,\pathRefs}
    \ExplSyntaxOn }
}
\ExplSyntaxOff

%%% ==================
%%% APPENDIX

% new page -- remove balance
\clearpage
\ExplSyntaxOn
\str_case_x:nn {\g_format_str} {
  {plain} {                     % ========== PLAIN
    \ExplSyntaxOff
    \ExplSyntaxOn }
  {hpec} {                      % ========== HPEC
    \ExplSyntaxOff
    \nobalance
    \ExplSyntaxOn }
}
\ExplSyntaxOff

% \appendix

% \input{tex/BODY-appendix}

\end{document}